\def\year{2021}\relax
\documentclass[letterpaper]{article} % DO NOT CHANGE THIS
\usepackage{aaai21}  % DO NOT CHANGE THIS
\usepackage{times}  % DO NOT CHANGE THIS
\usepackage{helvet} % DO NOT CHANGE THIS
\usepackage{courier}  % DO NOT CHANGE THIS
\usepackage[hyphens]{url}  % DO NOT CHANGE THIS
\usepackage{graphicx} % DO NOT CHANGE THIS
\usepackage{natbib}  % DO NOT CHANGE THIS AND DO NOT ADD ANY OPTIONS TO IT
\usepackage{caption} % DO NOT CHANGE THIS AND DO NOT ADD ANY OPTIONS TO IT
\usepackage{amsmath}
\usepackage{subfigure}
\usepackage{booktabs}
\usepackage{color}
\usepackage[switch]{lineno}  
\nocopyright

\title{Split then Refine: Stacked Attention-guided ResUNets for \\ Blind Single Image Visible Watermark Removal}

\author {
        Xiaodong Cun and
        Chi-Man Pun\thanks{Corresponding Author} \\
}
\affiliations {
    University of Macau, Macau, China \\
    yb87432@umac.mo, cmpun@umac.mo \\
}

\begin{document}
% \linenumbers  
\maketitle

\begin{abstract}
Digital watermark is a commonly used technique to protect the copyright of medias. Simultaneously, to increase the robustness of watermark, attacking technique, such as watermark removal, also gets the attention from the community. 
Previous watermark removal methods require to gain the watermark location from users or train a multi-task network to recover the background indiscriminately. 
However, when jointly learning, the network performs better on watermark detection than recovering the texture. 
Inspired by this observation and to erase the visible watermarks blindly, we propose a novel two-stage framework with a stacked attention-guided ResUNets to simulate the process of detection, removal and refinement. 
In the first stage, we design a multi-task network called SplitNet. It learns the basis features for three sub tasks altogether while the task-specific features separately use multiple channel attentions.
Then, with the predicted mask and coarser restored image, we design RefineNet to smooth the watermarked region with a mask-guided spatial attention.
Besides network structure, the proposed algorithm also combines multiple perceptual losses for better quality both visually and numerically. 
We extensively evaluate our algorithm over four different datasets under various settings and the experiments show that our approach outperforms other state-of-the-art methods by a large margin. The code is available at: \url{http://github.com/vinthony/deep-blind-watermark-removal}.
\end{abstract}

\begin{figure}[t]
\centering
  \includegraphics[width=\columnwidth]{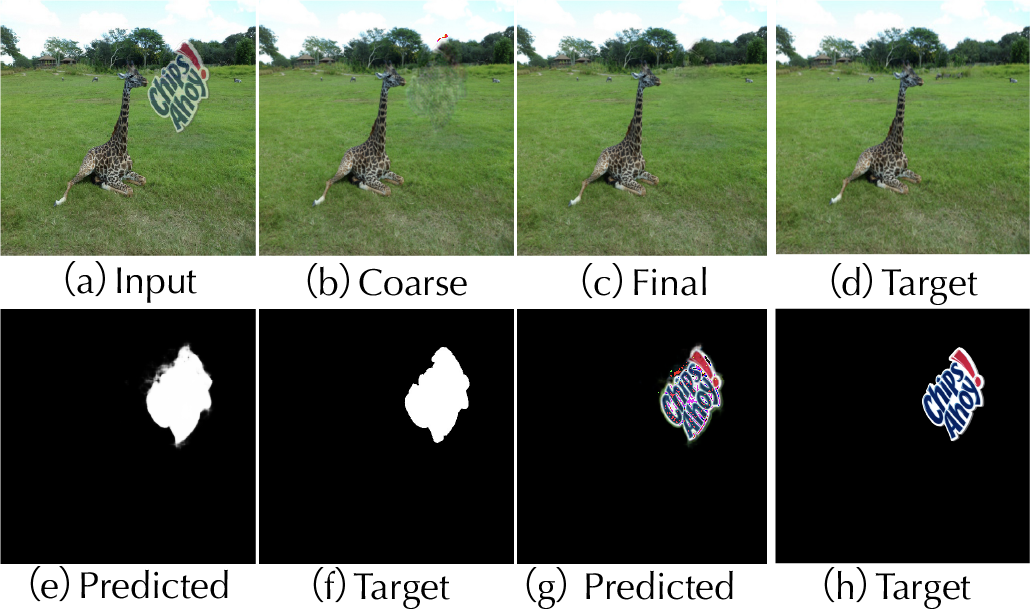}
  \caption{We propose an end-to-end, two-stage method to remove the visible watermark from a single image without any prior knowledge or user input. In the first stage, we use a multi-task attention-guided network to predict the coarser background~(b), the location of watermark~(e), and the recovered watermark~(g). Then, the second-stage network uses (b) and (e) as input and learns the final results~(c). }
  \label{fig:teaser}
\end{figure}

\section{Introduction}
Sharing rich contents such as images, audios or videos on social media has become a significant trend recently. 
Thus, digital watermarks, especially the visible ones, are used to credit the affiliation of the digital media. 
In general, the media vendor or the users assume the embedded watermarks are robust to various attacks, such as JPEG compression and the removal techniques. 
Also, the watermarks should have minimal visual influences on recognizing the original media contents. 
The robustness of the watermarking systems is a very important multimedia security issue for protecting the copyright or ownership of the digital media. Therefore, many research work have been studying the watermark removal methods to verify and improve the resilience of digital watermarks. 
%As we all know, keeping the robustness of watermarks is important to prevent the illegal usage of the media, such as prohibited business activities. In this paper, in order to improve the robustness of watermarks and get the attention from the community about the security issues in the visible watermark, we start from the opposite view of the multimedia security by attacking~(removing) the watermark using deep learning. 
%focus on dealing with the embedded visible watermarks in images and propose a novel method to attack this kind of watermark. 
%Our method is totally blind, which means it only need a single watermarked image as input without any prior knowledge.
%We also aim to get the attention from the community about security issues in the visible watermark. 
%become a significant trend recently  
%have/share the same meaning  ones   business occasion 
%also, minimal visual influence guarantees the recognization of 

Since the invention of the watermark, removal techniques~\cite{huang2004attacking,park2012identigram,pei2006novel} have been developed to attack them synchronously. 
In some previous work, detecting the location is a precondition for removing because it is relatively easier to match the features with the known mask. 
After getting the location, the watermark can be removed by image in-painting~\cite{huang2004attacking} or features matching~\cite{park2012identigram,pei2006novel}. 
On the other hand, since the single image-based watermark removal is difficult, multiple images with the same watermarks~\cite{dekel2017effectiveness,gandelsman2019double} supply a strong prior knowledge for removing.
However, erasing the watermark interactively and using multiple images restrict the application range of previous methods. 
%requiring the mask from user or using the multiple images restricts the application range of previous methods. 
Recently, \citet{hertz2019blind} use a multi-task network to remove the watermark with a single image blindly for the first time. 
Nevertheless, as discussed in the previous work~\cite{liu2018image,s2am,ulyanov2018deep}, the watermarked regions should be learnt explicitly using the carefully designed blocks. 
% and the detection is easier than recovery
%Besides, the training process of their network is unstable. 
Thus, we propose a novel method to tackle these issues and follow it for further study.

%our first observation is that three tasks(removing, getting the location and recovering the content of watermark) only....

The basic idea behind the proposed two-stage framework is: \textit{Considering the framework of multi-task learning, watermark removal is more complicated than detection due to the texture non-harmony. Thus, further refinement is necessary}. 
As shown in Fig.~\ref{fig:teaser}, if the background~(b), predicted mask~(e) and watermark~(g) are generated from a single network, the background~(b) performs worse compared with other tasks. It might be because the watermark removal should restore the exactly pixel values from the degraded region while the watermark detection only need to gain the binary mask. Consequently, although a single multi-task network can be used for watermark removal as in \citet{hertz2019blind}, it is still necessary to smooth the watermarked region, and we refine it with another network using the predicted mask~(e). 
%poor pool

To model the analysis above, in this paper, we propose SplitNet and RefineNet, a novel attention-guided two-stage framework using Residual Block-based UNet~(ResUNet) ~\cite{hertz2019blind} as the backbone of each stage.
In SplitNet, we jointly learn three tasks~(watermark removal, detection and recovery) using a single encoder and multiple decoders as in \citet{hertz2019blind}.
Differently, we improve the performance of the original method with several modifications. 
Firstly, inspired by the recent works in multi-domain learning~\cite{wang2019towards}, we share the parameters in all three decoders while learning the bias for each task separately using channel attentions~\cite{Hu:2017tf}.
In the second stage, we propose RefineNet to further refine the masked region pixels with the predicted mask and coarser results in SplitNet. 
Moreover, spatial-separated attention module~\cite{s2am} is involved into RefineNet and learns the masked region specifically. 
Besides two stage framework, the simple pixel-wise $L_{1}$ loss in \citep{hertz2019blind} might also produce the blur results in the restored region. Thus, similar to the image super-resolution~\cite{Johnson:2016wm} and reflection removal task~\cite{zhang2018perceptual}, we use the deep perceptual loss~\cite{Johnson:2016wm} and SSIM loss~\cite{wang2004image} for better visual and numerical quality. 
The main contributions of the paper are as follows:
\begin{itemize}
  \item We consider the task differences in multi-task watermark removal for the first time and formulate it as a two-stage framework by prediction and harmonization.
   
  \item In SplitNet, we regard the joint learning framework as a multi-domain learning problem and propose an accurate and compact model by domain~(task)-specific attention. 

  \item In RefineNet, we use the predicted mask and involve the mask-guided spatial attention modules for further harmonize the predicted regions. 

  \item The results show that our approach outperforms various state-of-the-art methods by a large margin.
  
\end{itemize}

\section{Related Work}

\textbf{Visible Watermark Removal}
Digital watermarks play an important role in commercial digital copyright protection. 
Most previous watermark removal methods need to indicate the location of the watermark with the user before removing it by hand-crafted features, such as independent component analysis~\cite{pei2006novel} and color space transformation~\cite{park2012identigram}. 
Besides, multiple images based visible watermark removal has also been widely investigated~\cite{dekel2017effectiveness,gandelsman2019double}. However, these methods require more prior knowledge, and removing the watermark by specific features only works on limited samples.

More recently, deep learning-based methods show great power in many computer vision tasks~\cite{he2016deep}, which has also been used to remove visible watermarks~\cite{li2019towards,cheng2018large, hertz2019blind}. 
However, a pre-trained watermark detector is necessary beforehand~\cite{cheng2018large}, or they only consider watermark removal as an image-to-image translation problem~\cite{li2019towards} as pix2pix~\cite{Isola2017}.
Besides, their dataset only contains gray-scale watermarks. 
Nevertheless, colorful images are more general nowadays. More recently, \citet{hertz2019blind} design a novel deep learning-based method to detect, remove, and recover the motif in a single forward.
However, their method still pays little attention to learning the degraded region specifically.

\textbf{Image In-painting} 
After identifying the specific watermark location by users, image in-painting techniques can also be used to remove the visible watermarks~\cite{huang2004attacking}. Thus, attention-guided in-painting methods, such as Partial Convolution~\cite{liu2018image}, Contextual Attention~\cite{yu2018generative} and Gated Convolution~\cite{Yu:2018uw} may also serve to watermark removal.
However, in-painting based watermark removal methods are supposed to point out the watermark location in the image by users. 
%After that, the watermark can be removed by the convolutional neural network. 
Moreover, in-painting based methods completely erase the masked region, and the background context will be re-generated by the neural network other than re-using the texture from the input.
Differently, in the task of watermark removal, the transparent watermark contains both background and foreground information. 
Ignoring background cures will destroy the original structure, gaining undesired results. More importantly, due to the highly ill-pose of image in-painting, current methods only work on limited types of images, such as faces. Differently, our method can work in the wild with the help of transparent information.

\begin{figure*}[t]
\centering
  \includegraphics[width=0.95\textwidth]{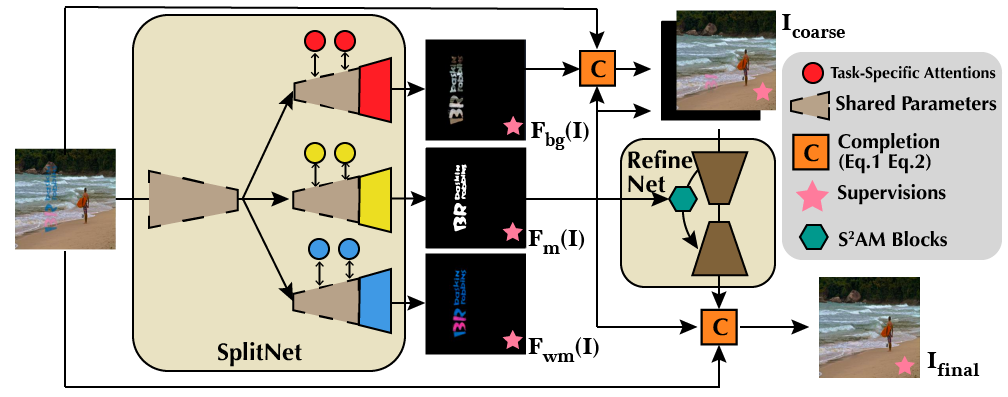}
  \caption{Proposed two-stage framework. We propose SplitNet to gain the coarser results by learning the watermark detection, removal and recovery jointly. Then, we propose RefineNet to smooth the learned region with the predicted mask and the recovered background from the previous stage. Thus, our network can be trained in an end-to-end fashion without any manual intervention. Note that, for clearness, we do not show any skip-connections between all the encoders and decoders.}
  \label{fig:nn}
\end{figure*}

\textbf{Image Restoration} 
Our task can also be considered as image restoration which only needs to restore certain regions. Thus, Blind shadow removal~\cite{wang2018stacked,Ding_2019_ICCV,cun2019ghostfree}, blind image harmonization~\cite{s2am} and reflection separation~\cite{zhang2018,yang2018seeing} share a similar goal with our task.
However, the related tasks have limited resources compared with ours. For example, in blind shadow removal, the network only gets supervisions from the shadow-free image and the corresponding shadow mask, and it is impossible to get the ground truth shadow for learning jointly.
Reflection removal/separation can generate paired reflection and transmission images from the dataset while it is hard to get the reflected region~(mask) as supervision.
Luckily, more information can be easily collected and used in our task, such as watermark location and content.

\section{Proposed Method}
\label{sec:method}
We regard the blind single image-based visible watermark removal as a two-stage task. 
As shown in Fig.~\ref{fig:nn}, in the first stage, given a single watermarked image $I$, we propose SplitNet $F$, a multi-task ResUNet inspired by multi-domain learning, to generate the coarser intermediate results: the recovered background image $F_{bg}(I)$, the location~(mask) of watermark $F_{m}(I)$ and the recovered watermark $F_{wm}(I)$. Thus, the coarser restored image $I_{coarse}$ can be written as: 
\begin{equation}
	I_{coarse} = F_{bg}(I) \times F_{m}(I) + I \times ( 1 - F_{m}(I)) 
	\label{eq:1}
\end{equation} 

As discussed previously, due to the difficulty of the tasks are different, further refinement is necessary for watermark removal. Thus, we propose RefineNet $R$ as the second stage, which uses $I_{coarse}$ and $F_{m}(I)$ to generate the final result $I_{final}$ and this network smooths the predicted watermarked region using a spatial attention mechanism. Finally, the refined result $I_{final}$ can be formulated by the predicted mask $F_{m}(I)$ and the original input $I$:
\begin{equation}
	I_{final} = R(I_{coarse}, F_{m}(I)) \times F_{m}(I) + I \times ( 1 - F_{m}(I)) 
\label{eq:2}
\end{equation}

Note that, although the proposed method is a cascaded framework, the inputs of the second network are totally generated by the output of the first stage. 
Thus, our network can be trained and evaluated in an end-to-end fashion without any manual intervention. 
Below, we give the details of the proposed SplitNet, RefineNet and the loss functions.

\subsection{SplitNet}
It has been widely testified in various computer vision tasks~\cite{liu2019end,ruder2017overview,wang2009joint} that learning multiple related targets jointly can boost the performance of a single task. 
Thus, we propose SplitNet, a multi-task network jointly learn the watermark, background and mask~(as shown in Fig.~\ref{fig:nn}) in the first stage. 
Similar to \citep{hertz2019blind}, our network uses the residual block based UNet~\cite{Isola2017,hertz2019blind,he2016deep}~(ResUNet) as the basic encoder-decoder structure. Specifically, we build a shared encoder and multiple decoders as the main structure and each encoder/decoder contains 5 different scales of stacked~(3 in each scale) ResBlocks to capture the multi-scale features. Getting benefits from the locally skip-connections in ResBlocks and globally skip-connections between the encoder and decoder, this backbone shows superior performance.

% Furthermore, we observe that the proposed network only need to learn the \textit{same} spatial region~(the outputs of SplitNet in Fig.~\ref{fig:nn}) in this three tasks. 
% The only difference between three outputs is the contents in this region. Thus, we design a novel network structure to achieve this observation. 

%However, we find the training process unstable if we directly deploy the network structure~\cite{hertz2019blind}.
%Thus, we make several improvements for better stability. As shown in Fig.~\ref{fig:res}, we illustrate a single layer of the proposed improved ResBlock in encoder and decoder. 
%Specifically, we replace all the batch normalization with instance normalization~\cite{ulyanov2016instance} because our task is more similar to the style transformation task in the specific region, and the normalization should be applied for each sample.
%Then, we concatenate all the features after the non-linear activation other than the mixture of convolutional feature and non-linear activation in the previous work~\cite{hertz2019blind}. 
%In the decoder, we also replace the original ReLU with LeakyReLU~\cite{xu2015empirical}, which is similar to UNet in ~\cite{Isola2017}. The proposed structure can hugely improve network stability and performance. More differences and comparisons can be found in supplementary material and experiments.

\begin{figure}[!t]
  \includegraphics[width=\columnwidth]{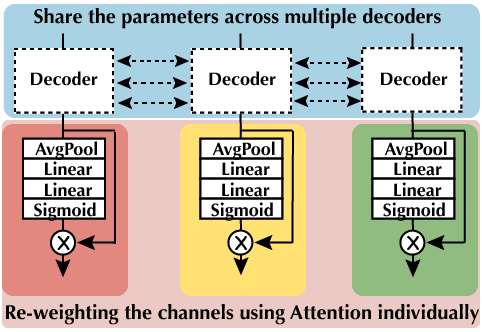}
  \caption{The detailed task-specific decoders in SplitNet. Our method shares the parameters among different tasks in the main stream and learns the features for each task individually using task-specific attentions.
  }
  \label{fig:decoder}
\end{figure}

Different from previous work, we consider the joint learning framework as a multi-domain learning problem~\cite{xiao2020multi,wang2019towards,liu2019end} in this task for the first time. 
In multi-domain learning, to learn an efficient model, almost all the parameters are shared in the training process while each domain needs to be emphasized using different parameters. 
Similarly, the three tasks in our framework focus on learning \textit{one spatial region} and each task has to learn its specific features for individual reconstruction. 
In detail, we share the learn-able parameters in all three high-level decoders \textit{altogether} and learn the specific feature for each task by domain attentions \textit{individually}. This strategy helps us build a more efficient and effective model. 
To model this observation, inspired by SE-Net~\cite{Hu:2017tf}, we design the task-specific attentions to re-weight the importance of the channels for each decoder~(task). 
Fig.~\ref{fig:decoder} gives a close look at the proposed framework in the decoder over multiple shared branches. 
In each level of the decoders, we learn the basis features for all three tasks using the ResBlocks. 
After that, these basis features are re-weighted by the task-specific attentions. 
The detailed structure of the task-specific attention is also illustrated in the Fig.~\ref{fig:decoder}.

Moreover, we optimize the structure of the original ResBlock in \citet{hertz2019blind} for stable training. Since it is not the main contribution of our paper, we give a more detailed explanation in the supplementary material.

\subsection{RefineNet}
The quality of the restored region remains poor if we predict the watermark-free image and mask from a single multi-task network~(as shown in Fig.~\ref{fig:teaser}).
However, the predicted mask achieves satisfying result, which might be because the texture recovery is far more difficult than detection. 
Therefore, we propose RefineNet for further refinement using the predicted coarse background and the location of watermark~(mask) from SplitNet. 

As discussed in previous works~\cite{liu2018image,ulyanov2018deep}, if we directly feed the mask and coarser results to UNet, the naive network might not focus on learning the masked region. 
Thus, inspired by the recently proposed Spatial-Separated Attention Module~(S$^2$AM~\cite{s2am}) for image harmonization, we refine the predicted masked regions through the attention-guided network. 
%taking ground truth mask as an input is necessary for the originalxxx
%as a part of input, ground truth mask should be xxxxx
Nevertheless, as a part of input, taking the mask from user is necessary in the original S$^2$AM.
%the vanilla S$^2$AM require the the mask from user as input.
Hence, we regard the predicted mask from SplitNet as a trustworthy label, feeding it into the RefineNet as both additional input channel and the mask of S$^2$AM block~(as shown in Fig.~\ref{fig:nn}). 
Moreover, since the ResBlock based network has been proven to capture the low-level feature well \cite{hertz2019blind}, we replace the original eight layers UNet in S$^2$AM with the proposed five layers ResBlock-based backbone as SplitNet. Then, the S$^2$AM is inserted in the two coarser levels of the decoder. Below, we give a short review of S$^2$AM and details of the overall network structure.

\begin{figure}[hbt]
  \includegraphics[width=\columnwidth]{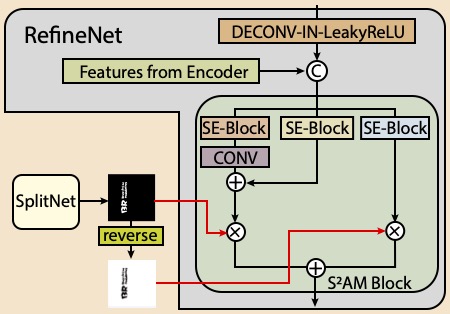}
  \caption{The predicted mask from the SplitNet is used in the S$^2$AM Block to learn the differences in the masked and non-masked regions respectively.}
  \label{fig:s2am}
\end{figure}

S$^2$AM is initially proposed in image harmonization to match the low-level features between the masked foreground and non-masked background. As shown in the Fig.~\ref{fig:s2am}, the S$^2$AM block is inserted after the concatenation between the features from encoder and decoder. Then, with the help of the hard-coded mask, three channel attentions~(SE-Block~\cite{Hu:2017tf}) are applied to re-weight the original features automatically. More details of S$^2$AM can be found in \cite{s2am}.

%\subsection{Network Structure Discussion}

%Notice that, our main contribution is the proposed two-stage framework, which means the ResUNets~\cite{hertz2019blind} is only one of the choices to build the networks. Other similar backbones can also be used, such as UNet~\cite{Isola2017} and context aggregation networks~\cite{zhang2018perceptual}. We choose ResUNet because it focuses on learning low-level details and gains better performance.  

\subsection{Loss Functions}
The proposed framework is supposed to optimize multiple targets at the same time. The predicted results include three outputs from SplitNet: the predicted background $I_{coarse}$, the mask $M'$ of watermark location, and the reconstructed watermark $I_{wm}$. Also, the final prediction $I_{final}$ comes from RefineNet. 
Below, we introduce the details in each part of the loss functions.

\subsubsection{Losses for Watermark Removal}
%For this reason, learning the watermark region is enough for recovery. 
Optimizing the losses of watermark removal is the main target of our framework, which contains multiple parts. Firstly, following the previous work~\cite{hertz2019blind}, we use the relative L1 loss over the ground truth mask $M$ by $\ell^{gt}_{r} = \frac{1}{sum(M)}|| M \times F_{bg}(I) - M \times I_{gt} ||_1 $ to learn the masked region particularly, where $F_{bg}(I)$ is the direct output for background prediction. As suggested by ~\citet{hertz2019blind}, calculating the loss on $M$ only helps to prevent the early over-fitting. Moreover, we add an additional loss over the predicted mask $\ell^{pred}_{r} = \frac{1}{sum(M)}|| M' \times F_{bg}(I) - M \times I_{gt} ||_1$ to squeeze the gap between the prediction and ground truth. Because when testing, we need to recover the image using the predicted mask~(as in Eq.~\ref{eq:1}, Eq.~\ref{eq:2}).

Besides, we also measure the quality of the recovered images $I_{coarse}$ and $I_{final}$. Computing the losses over the full image enables us to interpolate the frequently used perception losses into our system for better visual quality, such as deep perception loss~$\Phi_{vgg}$~\cite{zhang2018perceptual,Johnson:2016wm} and SSIM loss~$\Psi_{ssim}$~\cite{zhao2016loss,wang2004image}. 
In deep perceptual loss~\cite{zhang2018perceptual}, we extract the features between target and the predicted image from the pre-trained VGG16~\cite{simonyan2014very} in the layers of CONV1\_2, CONV2\_2,CONV3\_3, and then measure the $L_1$ distance in the feature domain. Finally, the total loss in the coarser stage $\ell_{coarse}$ and the final stage $\ell_{final}$ can be written as:
\begin{equation}
\begin{aligned}
&\ell_{x} =  \alpha\sum_{k\in1,2,3}|| \Phi_{vgg}^{k\_k}(I_{x}) - \Phi_{vgg}^{k\_k}(I_{gt})||_1 + \ell^{gt}_{r}\\ 
&+ \beta  \Phi_{ssim}(I_{x},I_{gt}))  + || I_{x} - I_{gt} ||_1 +  \ell^{pred}_{r},
\end{aligned}  
\label{eq:3}
\end{equation}
where the recovered images $ I_x (x \in \{coarse,final\})$ are  from SplitNet and RefineNet by Eq.~\ref{eq:1} and Eq.~\ref{eq:2}. We set all the $\alpha=0.025$ and $ \beta=0.15$ in $\ell_{coarse}$ and $\ell_{refine}$.

\subsubsection{Losses for Watermark Detection} Similar to salient object and shadow detection, detecting the watermarked region is a binary pixel-level segmentation. Thus, we choose Binary Cross Entropy as the loss function, which can be written as:
\begin{equation}
 \ell_{m} =  M\log(M')+(1-M)\log(1-M'),
\end{equation}
where $M$ and $M'$ are the ground truth mask and the predicted mask, respectively.

\subsubsection{Losses for Watermark Recovery} Because the target watermark is also available in the proposed dataset, we measure the restored watermark with the original one over the masked region as $\ell_{wm}$. Following the relative L1 losses $\ell^{pred}_{r}$ and $\ell^{gt}_{r}$ for watermark removal, $\ell_{wm}$ is defined as:
$\ell_{wm} = \ell^{pred''}_{r} + \ell^{gt''}_{r}$,
where $\ell^{pred''}_{r}$ and $\ell^{gt''}_{r}$ represent the relative L1 loss on the restored watermark $F_{wm}(I)$.

Overall, the total loss $\ell_{all}$ in our algorithm is a combination of the losses above:
\begin{equation}
 \ell_{all} = \ell_{coarse} + \ell_{refine} +  \ell_{wm} +  \ell_{m}.
\end{equation}

\begin{table*}[t]
\centering
\small
\caption{Comparisons between the proposed method and other state-of-the-art methods on all synthesized datasets.}
\label{tab:stoa}
\begin{tabular}{@{}l|lll|lll|lll|lll@{}}
\toprule
 & \multicolumn{3}{c|}{LOGO-H} & \multicolumn{3}{c|}{LOGO-L} & \multicolumn{3}{c|}{LOGO-Gray} & \multicolumn{3}{c}{LOGO-30K} \\ \hline
Metrics & PSNR & SSIM & LPIPS & PSNR & SSIM & LPIPS & PSNR & SSIM  & LPIPS & PSNR & SSIM  & LPIPS  \\ \hline

Original  &  28.33 & 0.9579 & 0.0748 & 35.12 &  0.9830 & 
0.0312 & 31.82 & 0.9730 & 0.0411 & 31.01 &  0.9698  & 0.0555 \\
UNet  &  30.51  &  0.9612 &  0.0544 & 34.87 &  0.9814 & 0.0297                      & 32.15 & 0.9728 & 0.0353 & 33.27  &  0.9744  &  0.0365 \\
SIRF  &  32.35  &   0.9673  & 0.0801  &   36.25  &  0.9825 &                    0.0655 & 34.33 & 0.9782 & 0.0672 &  34.63    &   0.9783   &  0.0718    \\
BS$^2$AM & 31.93 &  0.9677 & 0.0445 &  36.11 & 0.9839     &                    0.0223  & 32.91 & 0.9754 & 0.0305 &  34.88  &  0.9793  &  0.0283 \\
DHAN &   35.68 &  0.9809  &  0.0661 & 38.54  &  0.9887 &  0.0591 & 36.39 & 0.9836 & 0.0594 &  37.67  &  0.9860   &   0.0624 \\
BVMR  &  36.51 &  0.9799 &   0.0237 & 40.24    &   0.9895  & 0.0126 & 38.90 & 0.9873 &  0.0115 & 38.28  &   0.9852  & 0.0181 \\
Ours  &  \textbf{40.05} & \textbf{0.9897}  & \textbf{0.0115} & \textbf{42.53} & \textbf{0.9924} & \textbf{0.0087} & \textbf{42.01} & \textbf{0.9928} & \textbf{0.0073} & \textbf{41.27} & \textbf{0.9910} &\textbf{0.0114} \\ \bottomrule
\end{tabular}
\end{table*}

\subsection{Dataset Acquisition}
Because there is no public available dataset containing all necessary information in our framework, we synthesize multiple datasets for different purposes to evaluate the proposed method\footnote{
We will open source all the synthesized datasets for further research and comparison.} and we argue that these datasets may be sufficient enough for real-world cases since these watermarks are also man-made by software.

In detail, we choose the background~(host) images from VAL2014 subset of the MSCOCO~\cite{mscoco} dataset, which is similar to the previous work on visual motif removal~\cite{hertz2019blind} and image harmonization~\cite{s2am}. 
Differently, to simulate the real-world watermarks, we collect over 1k different and famous logos from the Internet. Then, the watermarked samples are generated by natural images and the watermark~(logo) in different locations, semi-transparency and sizes randomly. Each training/testing sample contains the synthesized watermarked image, the original background, the watermark and the mask of the watermark for supervisions.
All the watermarks and the background images are non-overlapping in training and validation partition, showing the advantage of the algorithms in the unseen samples. 
We also analyze the distribution of the size, transparency, and class in the watermark to avoid the dataset bias. More details can be found in the supplementary material. Below, we give the detail settings of the four synthesized datasets:

\subsubsection{LOGO-L} We synthesize over 12K training and 2K test samples using 40\% of the images and logos, respectively. In the dataset, the transparency of the watermarks ranges from 35\% to 60\%. Also, the watermarks are resized to 35\% to 60\% of the width~(or height) of the host images. Thus, This dataset is an easier one because the watermark size is relatively small and the background is easily recognized through the watermarked area.
%watermark size is relatively small and the background... 
%the transparency of the watermarks range from 35\% to 60\%
%the watermark size accounts for 35\% to 60\% of the host image in each picture.

\subsubsection{LOGO-H} We create a harder sub-dataset in LOGO-H containing the same quantity of samples as LOGO-L. The watermark size in this dataset accounts for 60\% to 85\% of the host image. Besides, we also randomly set the transparency from 60\% to 85\%. Thus, the watermark in this dataset is difficult to be removed due to the missing texture and the larger degraded regions.

\subsubsection{LOGO-Gray} In real cases, the embedded watermark is usually a gray-scale image. Therefore, we create a LOGO-Gray sub-dataset to evaluate the performance which only contains gray-scale watermarks. This dataset also includes 12K images for training and 2K images for testing, as in LOGO-L and LOGO-H. The transparency and the size of the watermark are randomly chosen from 35\% to 85\%.

\subsubsection{LOGO30K} We create a larger dataset by synthesizing the watermarks with various size, location, and transparency. In LOGO30K, up to 28k and 4k images will be trained and tested respectively. The watermark size and the transparency range from 35\% to 85\%.

\newcommand{\xwidth}{0.115}
\newcommand{\xheight}{0.12}
\newcommand{\ima}[1]{figures/compare/000000334719-Dominos_Pizza_Logo-178/27kpng_#1}
\newcommand{\imf}[1]{figures/compare/000000136040-Kaporal_Jeans_Logo-96/10kgray_#1}
\newcommand{\ime}[1]{figures/compare/COCO_val2014_000000228290-Holland_Barrett_Logo-136/10kmid_#1}
\newcommand{\imi}[1]{figures/compare/000000458073-C_A_Logo-207/10khigh_#1}

\newcommand{\methods}{input_256,unet,urasc,sirfimage,gfsrimage,bvmr,vvv4n,target_256}

\begin{figure*}[!tb]
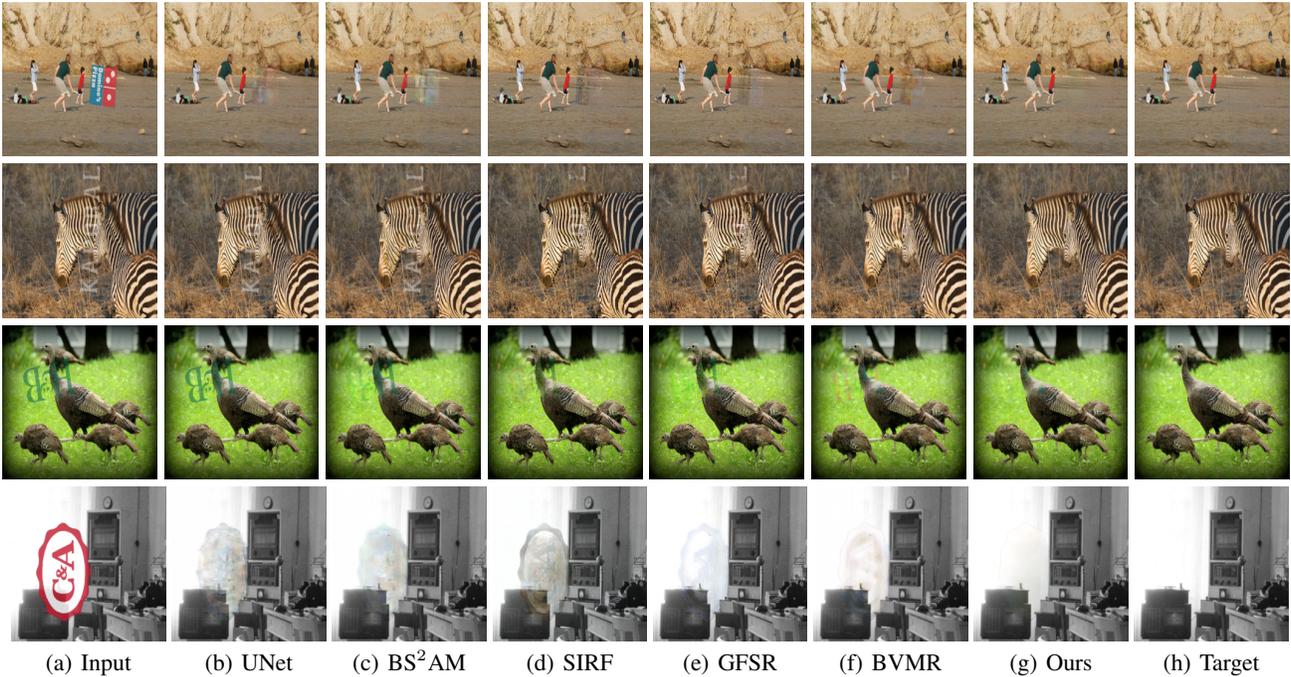

\centering     %%% not \center
\makeatletter
\@for\sun:=\methods\do{\subfigure{\centering\includegraphics[width=\xwidth\textwidth]{\ima{\sun}}}\hspace{0.3em}} % 27kpng
\par\bigskip\vspace*{-2.em}

\@for\sun:=\methods\do{\subfigure{\centering\includegraphics[width=\xwidth\textwidth]{\imf{\sun}}}\hspace{0.3em}} % 10kgray
\par\bigskip\vspace*{-2.em}

\@for\sun:=\methods\do{\subfigure{\centering\includegraphics[width=\xwidth\textwidth]{\ime{\sun}}}\hspace{0.3em}} % 27kpng
\par\bigskip\vspace*{-2.em}

\makeatother

\addtocounter{subfigure}{-24} %10khigh
\subfigure[Input]{\centering\includegraphics[width=\xwidth\textwidth]{\imi{input_256}}}
\subfigure[UNet]{\centering\includegraphics[width=\xwidth\textwidth]{\imi{unet}}}
\subfigure[BS$^2$AM]{\centering\includegraphics[width=\xwidth\textwidth]{\imi{urasc}}}
\subfigure[SIRF]{\centering\includegraphics[width=\xwidth\textwidth]{\imi{sirfimage}}}
\subfigure[GFSR]{\centering\includegraphics[width=\xwidth\textwidth]{\imi{gfsrimage}}}
\subfigure[BVMR]{\centering\includegraphics[width=\xwidth\textwidth]{\imi{bvmr}}}
\subfigure[Ours]{\centering\includegraphics[width=\xwidth\textwidth]{\imi{vvv4n}}}
\subfigure[Target]{\centering\includegraphics[width=\xwidth\textwidth]{\imi{target_256}}}

\caption{Comparison with state-of-the-art methods on the proposed four synthesized datasets. These four images are taken from LOGO-30K, LOGO-Gray, LOGO-L and LOGO-H from top to bottom.}

\label{fig:stoa}
\end{figure*}

\section{Experiments}
\textbf{Implementation details}
We use PyTorch~\cite{NEURIPS2019_9015} with CUDA v10.0 to implement our algorithm. To simplify the task and compare fairly, we conduct all the experiments on the image with a resolution of $256\times256$ and run 100 epochs for converging. The training batch size equals to 4, and all the optimizers are Adam~\cite{Kingma:2014us} with the learning rate of $1e^{-3}$. Following previous works~\cite{hertz2019blind,s2am}, we evaluate our approach with other state-of-the-art methods on the proposed four datasets using several popular numerical criteria, such as Structural Similarity~(SSIM~\cite{wang2004image}), Peak Signal-to-Noise Radio~(PSNR) and the deep perceptual similarity~(LPIPS\cite{zhang2018perceptual}).

\subsubsection{Comparisons with state-of-the-art methods}
Nowadays, few visible watermark removal methods are based on deep learning, expect for the most related blind visual motif removal method BVMR~\cite{hertz2019blind} and the deep watermark removal methods~\cite{cheng2018large,li2019towards} using the UNet-like~\cite{Isola2017} structures in image-to-image translation. Thus, we also compare our algorithm with some learning-based methods on related tasks. For example, blind image harmonization on a soft-masked version of the spatial-separated attention model~(BS$^2$AM\cite{s2am}); the learning-based reflection separation method~(SIRF~\cite{zhang2018perceptual}), which uses context aggregation network and perceptual loss; attention-guided dual hierarchical aggregation network for blind shadow removal~(DHAN~\cite{cun2019ghostfree}).  
As shown in Tab.~\ref{tab:stoa}, our algorithm achieves much better results in both shallow perceptual metrics~(PSNR,SSIM) and deep perceptual criterion~(LPIPS). The significant improvement on the harder sub-dataset LOGO-H verifies the assumption that a single model cannot work perfectly in the masked region removal.
Besides numeric metrics, our method also shows better visual quality than others. As shown in Fig.~\ref{fig:stoa},
naive UNet-based methods~(UNet, BS$^2$AM) fail when the watermark is large and the transparency is low. Both SIRF and GFSR can only remove certain parts of the watermark because their methods use limited supervisions. For instance, SIRF separates the layers from the watermark and background, and GFSR learns to remove and detect jointly.  Although BVMR outperforms other previous methods, it also shows noticeable artifacts due to the detection errors and texture misunderstanding. Differently, the proposed method uses all the information as supervisions and shows much better results. Interestingly, in some samples~(such as the top one in Fig.~\ref{fig:stoa}), our method cannot completely recover the color and texture of the background, while the images look more natural than other methods. This phenomenon also indicates the advantage of the proposed two-stage network with texture harmony in the masked region. We show more comparisons on the synthesized datasets and on ``real world'' samples~\cite{dekel2017effectiveness} in the supplementary material.

%\subsection{Ablation Studies}
\subsubsection{Ablation Studies}
\begin{figure}[!t]
\centering     %%% not \center
\includegraphics[width=\columnwidth]{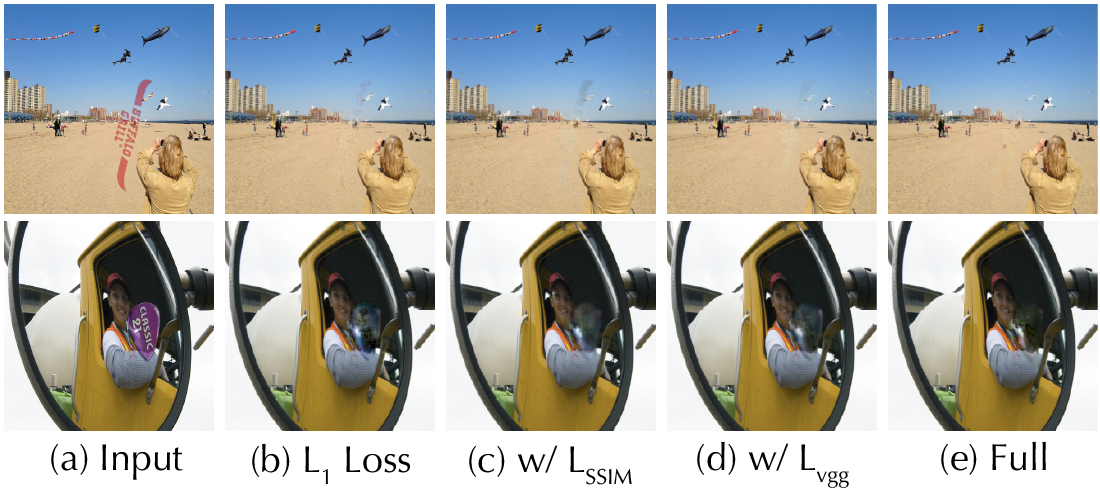}

\caption{The visual quality influence of loss functions in the final results. (b) means there is only relatively $\ell_{l1}$ losses for watermark removal, (c) and (d) refer to the additional $\ell_{ssim}$ and $\ell_{vgg}$, respectively. (e) is our full losses in Eq~\ref{eq:3}.}

\label{fig:loss}
\end{figure}

As shown in Tab.~\ref{tab:ab}, we evaluate the necessity of each component in our framework by removing or replacing it with other alternatives. We start the ablation study using the basic structure of ResUNet~\cite{hertz2019blind} as our SplitNet~(ResB). Then, we build the proposed framework by adding the naive ResUNet as RefineNet~(ResB). However, the naive RefineNet cannot improve the performance, especially on SSIM. It might be because the refinement is essential in the masked region. Thus, when we add the S$^2$AM module into the ResUNet(ResS$^2$AM), the RefineNet gains the knowledge from the predicted mask and learns to recover the masked region specifically. Besides, the proposed improved ResBlock~(iResB) and task-specific attentions~(iResBat) in the SplitNet also achieve significant progress. Apart from network structure, the additional perceptual losses also play a critical role in recovering the watermarked region details as shown in Fig.~\ref{fig:loss}, and we also give some numerical comparisons in Table~\ref{tab:ab} to support our claims on $\ell_{vgg}$ and $\ell_{ssim}$.

\begin{table}[t]
\small
\centering
\caption{Ablation study on the LOGO-Gray dataset.}

\label{tab:ab}
\begin{tabular}{llllll}
\toprule
SplitNet & RefineNet & additional loss     & PSNR & SSIM  \\ \hline
ResB  &           &       & 38.90 & 0.9873  \\
ResB  & ResB   &          & 39.34 & 0.9872 \\
ResB  & ResS$^2$AM   &    & 40.05 & 0.9895  \\
iResB  & iResS$^2$AM   &  & 41.77 & 0.9924 \\
%iResBat & ResS$^2$AM  &  & 41.69 & 0.9924  \\
iResBat & iResS$^2$AM  &  & 41.87 & 0.9925  \\
iResBat  & iResS$^2$AM  & $\ell_{ssim}$ & 41.89  &  0.9927       \\
iResBat  & iResS$^2$AM  & $\ell_{vgg} + \ell_{ssim}$ &  \textbf{42.01} & \textbf{0.9928}  \\\bottomrule
\end{tabular}
\end{table}

\subsubsection{Model Size}
The proposed framework contains two sub-networks, which naturally has more learning-able parameters than our baseline BVMR~\cite{hertz2019blind}. Consequently, the performance comparison on the fair model sizes is also necessary. As shown in Table~\ref{tab:mm}, we modify the original BVMR~(channel equals to 32 and depth equals to 5) to a larger network. For a fair comparison, all the networks are trained under the same loss function. Interestingly, the results of the deeper or heavier network structures in BVMR performs worse than the original one. It is not surprising that fewer parameters in BVMR lead to better results. On the one hand, it might be because the hyper-parameters of the network are the relatively best choice by network architecture searching. On the other hand, it might due to the unstable training in their network~(see more discussions in supplementary material). Differently, our network adds relatively few parameters and gains much better performance.

\subsubsection{Intermediate Results}
In Fig.~\ref{fig:teaser}, we have shown all the intermediate results in our framework for a better understanding of the proposed pipeline. It is clear that the coarse output from the SplitNet contains noticeable artifacts because SplitNet only includes a naive traditional convolution network. However, the final results in RefineNet perform better. It is also noticeable that although the predicted mask is not always completely correct in our network, it has minimal influence for the final prediction. More visualization results can be found in the supplementary material.

\begin{table}
\centering
\small
\caption{Comparison our method with BVMR. }
\label{tab:mm}
\begin{tabular}{l|c|cc}
\toprule
Methods & Parameters & PNSR & SSIM   \\ \hline
BVMR~(original) & \textbf{20.51M} & 38.90 & 0.9873  \\
BVMR~(channel=44) & 38.77M & 37.18 & 0.9832   \\
BVMR~(channel=48) & 46.14M & 37.78 & 0.9847   \\
BVMR~(depth=6) & 82.22M & 36.01 & 0.9812 \\
Ours &  32.62M & \textbf{42.01} & \textbf{0.9928}  \\ \bottomrule
\end{tabular}

\end{table}

\subsubsection{Limitation}
\begin{figure}[h]
  \includegraphics[width=\columnwidth]{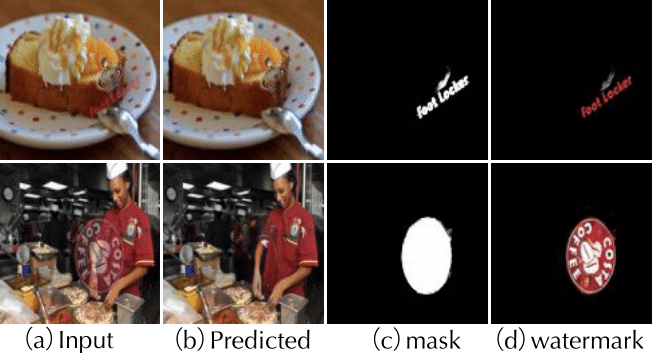}
  \caption{The limitation of the proposed network. Here, (b), (c) and (d) are the predicted results from our framework. }
  \label{fig:limit}
\end{figure}

Our method also suffer some limitations. As shown in the Fig.~\ref{fig:limit}, when the detection fails~(human in the logo cannot be detected) or the textures in the watermark and background are similar, our network cannot recover the watermark perfectly. However, these issues can often be ameliorated by a larger dataset or a stronger network. Another important limitation is the running speed, the proposed method runs 31fps on 256x256 images which is slower than baseline BVMR~(67fps), however, the accuracy is more meaningful than speed since our task is often used as a post-processing tool.

\section{Conclusion}
Following the observation that detection is much easier than removal, in this paper, we propose a novel two-stage framework, SplitNet and RefineNet, for blind single image-based visible watermark removal.
The SplitNet obtains the benefits from multi-task learning to generate the coarser outputs~(watermark, mask and background).
Also, in SplitNet, inspired by multi-domain learning, we build a compact network by sharing the parameters in the main stream decoders, while learning the task-specific attention individually.  
Then, the RefineNet utilizes the outputs from the previous stage and learns to refine the predicted region with spatial attention mechanism. 
Besides blind visual motif/watermark removal, our method could also be applied to other related tasks, such as blind image harmonization, shadow removal, and reflection removal in future work. 

\section{Acknowledgments}
This work was partly supported by the University of Macau under Grants: MYRG2018-00035-FST and MYRG2019-00086-FST, and the Science and Technology Development Fund,~Macau~SAR~(File~no.~0034/2019/AMJ, 0019/2019/A).

\bibliographystyle{aaai21}
\bibliography{wr.bib}

\clearpage
\begin{center}
\textbf{\large Supplemental Materials}
\end{center}

\begin{figure}[h]
\centering
  \includegraphics[width=0.8\columnwidth]{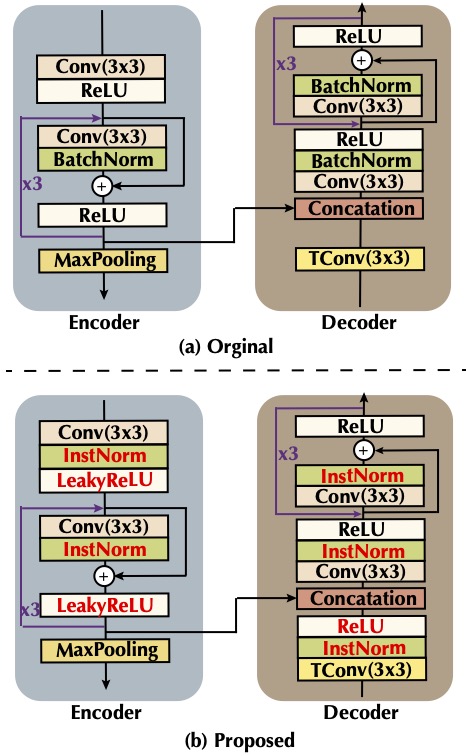}
  \caption{The differences between the proposed improved ResBlock and the original ResBlock. We mark the modification in red for easier recognization. Although we merely change few layers, our method achieves better and robuster results. }
  \label{fig:res}
\end{figure}

\begin{figure}[h]
  \includegraphics[width=0.9\columnwidth]{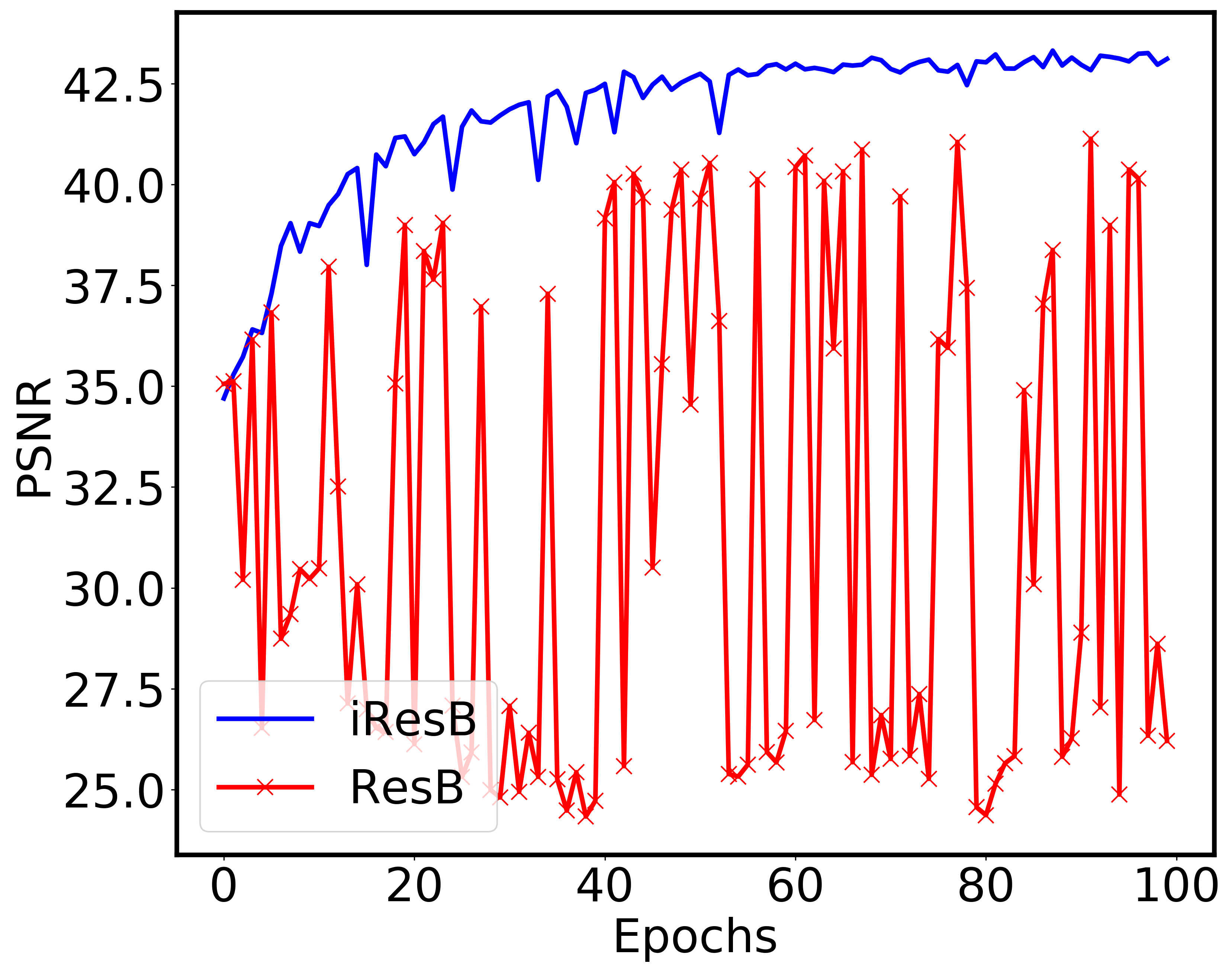}
  \caption{The PSNR of the ResBlock and the proposed iResBlock under the \textit{same} network structure on the validation set of LOGO-L. It is obvious that the proposed block is more stable in training.}
  \label{fig:stable}
\end{figure}

\subsubsection{Differences in ResBlocks}
We find the training process is unstable if we directly deploy the network structure~\cite{hertz2019blind}.
Thus, we make several improvements for better stability. As shown in Figure~\ref{fig:res}, we illustrate a single layer of the proposed improved ResBlock in the ResBlock based UNet comparing with the original. 
In detail, we replace all the batch normalization with instance normalization~\cite{ulyanov2016instance} because our task is more similar to the style transformation task in the specific region, and the normalization should be applied for each sample.
Then, we concatenate all the features after the non-linear activation other than the mixture of convolutional feature and non-linear activation in the previous work~\cite{hertz2019blind}. 
In the decoder, we also replace the original ReLU with LeakyReLU~\cite{xu2015empirical}, which is similar to UNet in ~\cite{Isola2017}. 

The proposed structure can hugely improve network stability and performance. As shown in Figure~\ref{fig:stable}, we plot the PSNR between the original ResBlock~(ResB) and the proposed block~(iResB) on the validation set of LOGO-L dataset. With the help of the carefully designed block, the proposed network can achieve better performance, and the volatility of the curve is more stable.

\subsubsection{Dataset Analysis}
\begin{figure}[h]
  \includegraphics[width=0.9\columnwidth]{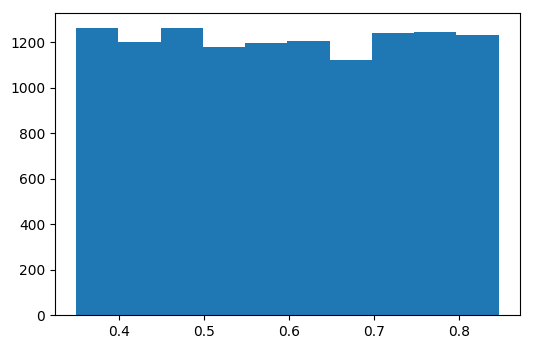}
  \vspace{-1em}
  \caption{The histogram of the distribution on watermark transparency. The percentage of the transparency are distributed equally among the dataset. }
  \label{fig:trans}
\end{figure}

\begin{figure}[h]
  \includegraphics[width=0.9\columnwidth]{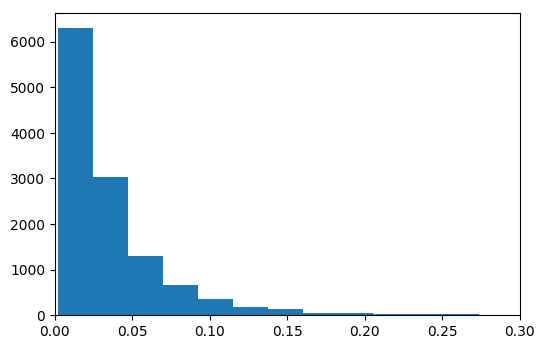}
    \vspace{-1em}
  \caption{We plot the distribution of the watermark area in the host image. Most watermarks are much smaller than the host image. This is in accord with the real world samples.}
   \label{fig:area}
\end{figure}

\begin{figure*}[hbt]
  \includegraphics[width=\textwidth]{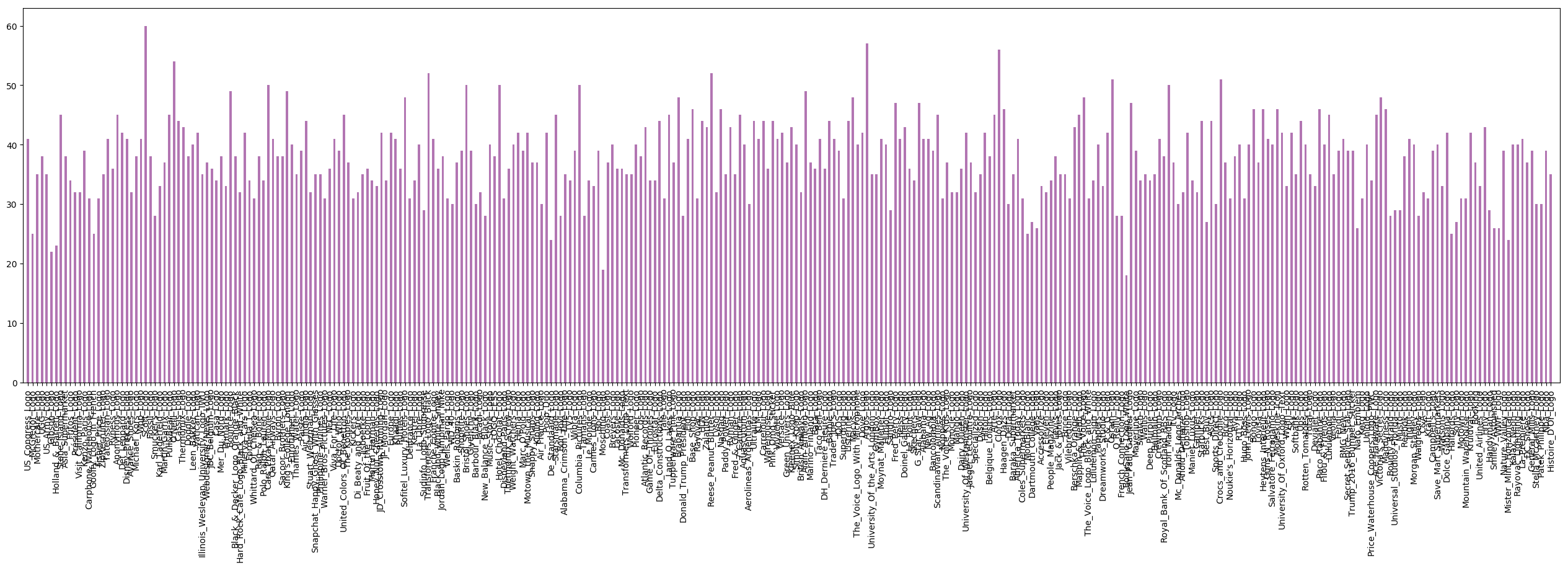}
  \vspace{-2em}
  \caption{The logo classes analysis in the training set of LOGO-Gray Dataset. The training subset of LOGO-Gray dataset contains 300+ different logos. Almost all the classes are equally distributed in the dataset.}
    \label{fig:classes}
\end{figure*}

Using the training set of LOGO-Gray dataset as an example, we analyze the distribution of the type, transparency, size of the watermark in the proposed dataset. In Figure~\ref{fig:classes}, the proposed dataset contains around 300 different classes~(logos) for training. Almost each class has similar samples in the overall dataset, which avoids the training bias. We also plot the distribution of the transparency over all samples~(as shown in Figure~\ref{fig:trans}). It shows that the transparency is also equally distributed. Moreover, we analyze the watermark size among this dataset as shown in Figure~\ref{fig:area}, our dataset contains more small watermarks which is similar to the real-world distribution.

\subsubsection{More results for comparison}
We plot more comparisons on the synthesized dataset to show the benefits of the proposed method as Figure~\ref{fig:inter} and Figure~\ref{fig:more}. Besides, we evaluate the baseline network~(BVMR) with ours on the ``real" dataset in \cite{dekel2017effectiveness}. This dataset only contains the input samples and we only present some comparisons visually. From Figure~\ref{fig:cvpr2017}, our network can remove and recover the watermark successfully. Notice that, these results are generated using the pre-trained model from LOGO-Gray dataset. This experiment shows that the proposed method generalizes well on the novel scenes.

\newcommand{\xxwidth}{0.15}
\newcommand{\xxheight}{0.15}

\newcommand{\xxa}[1]{figures/supp/xx/fotolia_3805308_#1}
\newcommand{\xxb}[1]{figures/supp/xx/fotolia_539921_#1}
\newcommand{\xxc}[1]{figures/supp/xx/AdobeStock_50850528_WM_#1}
\newcommand{\xxd}[1]{figures/supp/xx/AdobeStock_51093918_WM_#1}

\newcommand{\xxs}{input,coarse,final,mask,wm,vm3_final}

\begin{figure*}[!tb]
\centering     %%% not \center
\makeatletter
\@for\sun:=\xxs\do{\subfigure{\centering\includegraphics[width=\xxwidth\textwidth]{\xxa{\sun}}}\hspace{0.3em}}
\par\bigskip\vspace*{-1.7em}

\@for\sun:=\xxs\do{\subfigure{\centering\includegraphics[width=\xxwidth\textwidth]{\xxb{\sun}}}\hspace{0.3em}}
\par\bigskip\vspace*{-1.7em}

\@for\sun:=\xxs\do{\subfigure{\centering\includegraphics[width=\xxwidth\textwidth]{\xxc{\sun}}}\hspace{0.3em}}
\par\bigskip\vspace*{-1.7em}

\makeatother

\addtocounter{subfigure}{-18}
\subfigure[Input]{\centering\includegraphics[width=\xxwidth\textwidth]{\xxd{input}}}
\subfigure[Ours(Coarse)]{\centering\includegraphics[width=\xxwidth\textwidth]{\xxd{coarse}}}
\subfigure[Ours(Final)]{\centering\includegraphics[width=\xxwidth\textwidth]{\xxd{final}}}
\subfigure[Predicted Mask]{\centering\includegraphics[width=\xxwidth\textwidth]{\xxd{mask}}}
\subfigure[Predicted watermark]{\centering\includegraphics[width=\xxwidth\textwidth]{\xxd{wm}}}
\subfigure[BVMR~\cite{hertz2019blind}]{\centering\includegraphics[width=\xxwidth\textwidth]{\xxd{vm3_final}}}
\vspace{-1em}
\caption{More comparisons between our method and BVMR~\cite{hertz2019blind} over the dataset in \cite{dekel2017effectiveness}. }
\vspace{-1em}
\label{fig:cvpr2017}
\end{figure*}

\newcommand{\xa}[1]{figures/supp/inter/COCO_val2014_000000058647-Fun_Radio_Logo-107_#1}
\newcommand{\xb}[1]{figures/supp/inter/COCO_val2014_000000162366-Filou__Friends_Logo-172_#1}
\newcommand{\xc}[1]{figures/supp/inter/COCO_val2014_000000408859-Press_Shop_Logo-125_#1}
\newcommand{\xd}[1]{figures/supp/inter/COCO_val2014_000000246876-Stanford_Transparent_Logo-114_#1}
\newcommand{\xe}[1]{figures/supp/inter/COCO_val2014_000000332455-Chips_Ahoy_Logo-116_#1}
\newcommand{\xf}[1]{figures/supp/inter/COCO_val2014_000000501502-Scribbler_Logo-160_#1}
\newcommand{\xg}[1]{figures/supp/inter/COCO_val2014_000000394892-Burger_King_Logo-192_#1}
\newcommand{\xh}[1]{figures/supp/inter/COCO_val2014_000000402077-TV5_Logo-173_#1}
\newcommand{\xii}[1]{figures/supp/inter/COCO_val2014_000000426687-Eldi_Logo-151_#1}
\newcommand{\xj}[1]{figures/supp/inter/COCO_val2014_000000021905-Subway_Logo-130_#1}

\newcommand{\wwwws}{input,coarse,final,target,mask,gt_mask,wm,gt_wm}

\begin{figure*}[!tb]
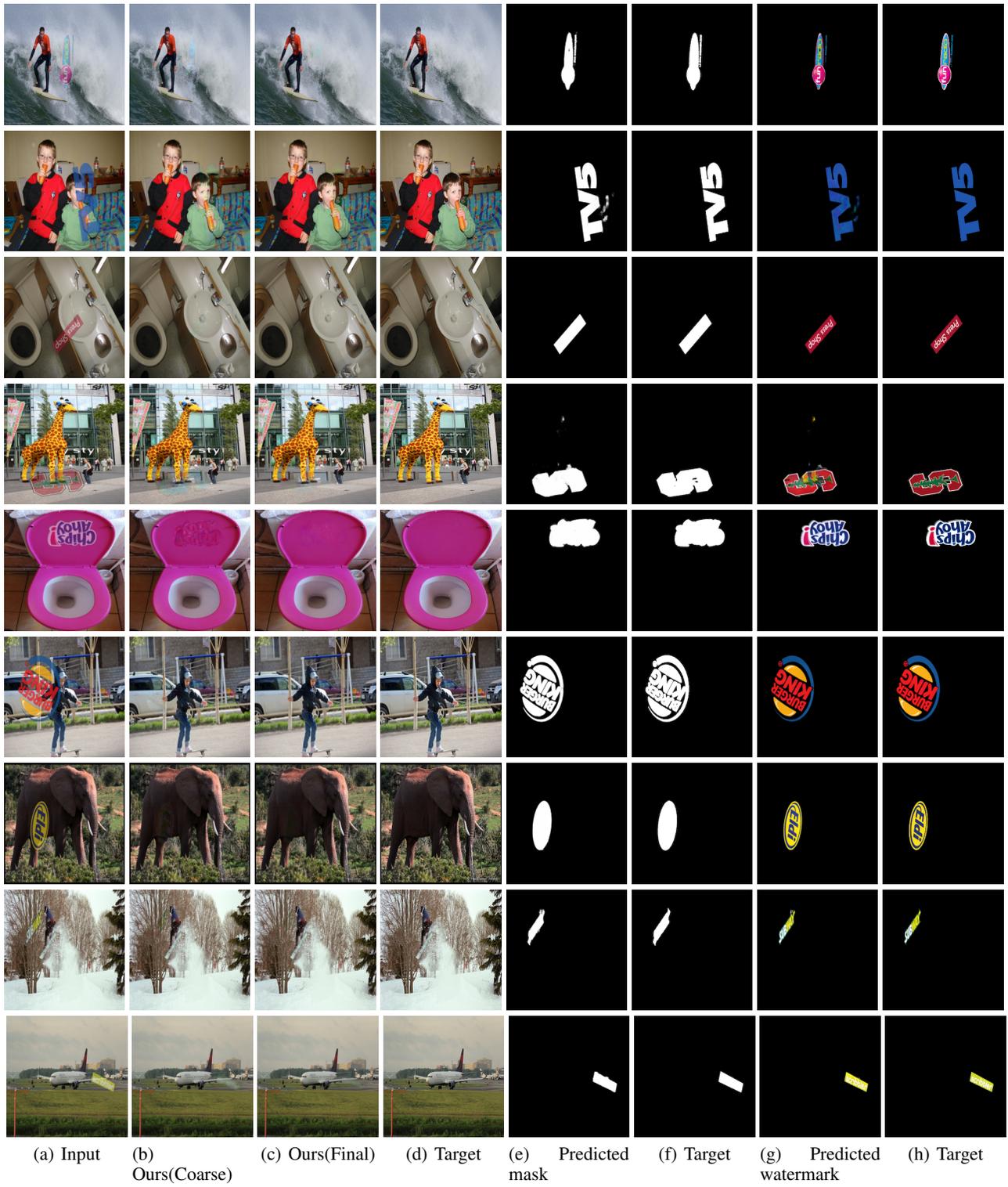

\centering     %%% not \center
\makeatletter
\@for\sun:=\wwwws\do{\subfigure{\centering\includegraphics[width=\xwidth\textwidth]{\xa{\sun}}}\hspace{0.25em}}
\par\bigskip\vspace*{-2.em}

\@for\sun:=\wwwws\do{\subfigure{\centering\includegraphics[width=\xwidth\textwidth]{\xh{\sun}}}\hspace{0.25em}}
\par\bigskip\vspace*{-2.em}

\@for\sun:=\wwwws\do{\subfigure{\centering\includegraphics[width=\xwidth\textwidth]{\xc{\sun}}}\hspace{0.25em}}
\par\bigskip\vspace*{-2.em}

\@for\sun:=\wwwws\do{\subfigure{\centering\includegraphics[width=\xwidth\textwidth]{\xd{\sun}}}\hspace{0.25em}}
\par\bigskip\vspace*{-2.em}

\@for\sun:=\wwwws\do{\subfigure{\centering\includegraphics[width=\xwidth\textwidth]{\xe{\sun}}}\hspace{0.25em}}
\par\bigskip\vspace*{-2.em}

\@for\sun:=\wwwws\do{\subfigure{\centering\includegraphics[width=\xwidth\textwidth]{\xg{\sun}}}\hspace{0.25em}}
\par\bigskip\vspace*{-2.em}

\@for\sun:=\wwwws\do{\subfigure{\centering\includegraphics[width=\xwidth\textwidth]{\xii{\sun}}}\hspace{0.25em}}
\par\bigskip\vspace*{-2.em}

\@for\sun:=\wwwws\do{\subfigure{\centering\includegraphics[width=\xwidth\textwidth]{\xj{\sun}}}\hspace{0.25em}}
\par\bigskip\vspace*{-2.em}
%
%\@for\sun:=\wwwws\do{\subfigure{\centering\includegraphics[width=\xwidth\textwidth]{\xf{\sun}}}\hspace{0.3em}}
%\par\bigskip\vspace*{-2.5em}

\makeatother

\addtocounter{subfigure}{-64}
\subfigure[Input]{\centering\includegraphics[width=\xwidth\textwidth]{\xf{input}}}
\subfigure[Ours(Coarse)]{\centering\includegraphics[width=\xwidth\textwidth]{\xf{coarse}}}
\subfigure[Ours(Final)]{\centering\includegraphics[width=\xwidth\textwidth]{\xf{final}}}
\subfigure[Target]{\centering\includegraphics[width=\xwidth\textwidth]{\xf{target}}}
\subfigure[Predicted mask]{\centering\includegraphics[width=\xwidth\textwidth]{\xf{Mask}}}
\subfigure[Target]{\centering\includegraphics[width=\xwidth\textwidth]{\xf{gt_mask}}}
\subfigure[Predicted watermark]{\centering\includegraphics[width=\xwidth\textwidth]{\xf{wm}}}
\subfigure[Target]{\centering\includegraphics[width=\xwidth\textwidth]{\xf{gt_wm}}}

\caption{More intermedia results of the proposed two-stage framework.}
\label{fig:inter}
\end{figure*}

\newcommand{\sima}[1]{figures/supp/000000034115-Sabian_Logo-165/10kgray_#1}
\newcommand{\simb}[1]{figures/supp/COCO_val2014_000000346587-JoJo_Maman_bebe_Logo-216/10khigh_#1_COCO_val2014_000000346587-JoJo_Maman_bebe_Logo-216}
\newcommand{\simc}[1]{figures/supp/COCO_val2014_000000113173-Burger_King_Logo-175/27kpng_#1_COCO_val2014_000000113173-Burger_King_Logo-175}
\newcommand{\sime}[1]{figures/supp/COCO_val2014_000000160239-History_Channel-141/27kpng_#1_COCO_val2014_000000160239-History_Channel-141}
\newcommand{\simf}[1]{figures/supp/COCO_val2014_000000268363-Snapchat_Happy_Ghost_With_Glasses-170/10khigh_#1_COCO_val2014_000000268363-Snapchat_Happy_Ghost_With_Glasses-170}
\newcommand{\simg}[1]{figures/supp/COCO_val2014_000000329336-Sci_Fi_Universal_Logo-170/10khigh_#1_COCO_val2014_000000329336-Sci_Fi_Universal_Logo-170}
\newcommand{\simh}[1]{figures/supp/COCO_val2014_000000353909-Kodak_Logo-215/10khigh_#1_COCO_val2014_000000353909-Kodak_Logo-215}
\newcommand{\simi}[1]{figures/supp/COCO_val2014_000000542089-Gamma_Logo-160/27kpng_#1_COCO_val2014_000000542089-Gamma_Logo-160}
\newcommand{\simk}[1]{figures/supp/COCO_val2014_000000356533-Classic_21_Logo-146/27kpng_#1_COCO_val2014_000000356533-Classic_21_Logo-146}

\newcommand{\smethods}{input_256,unet,urasc,sirfimage,gfsrimage,bvmr,vvv4n,target_256}

\begin{figure*}[!tb]
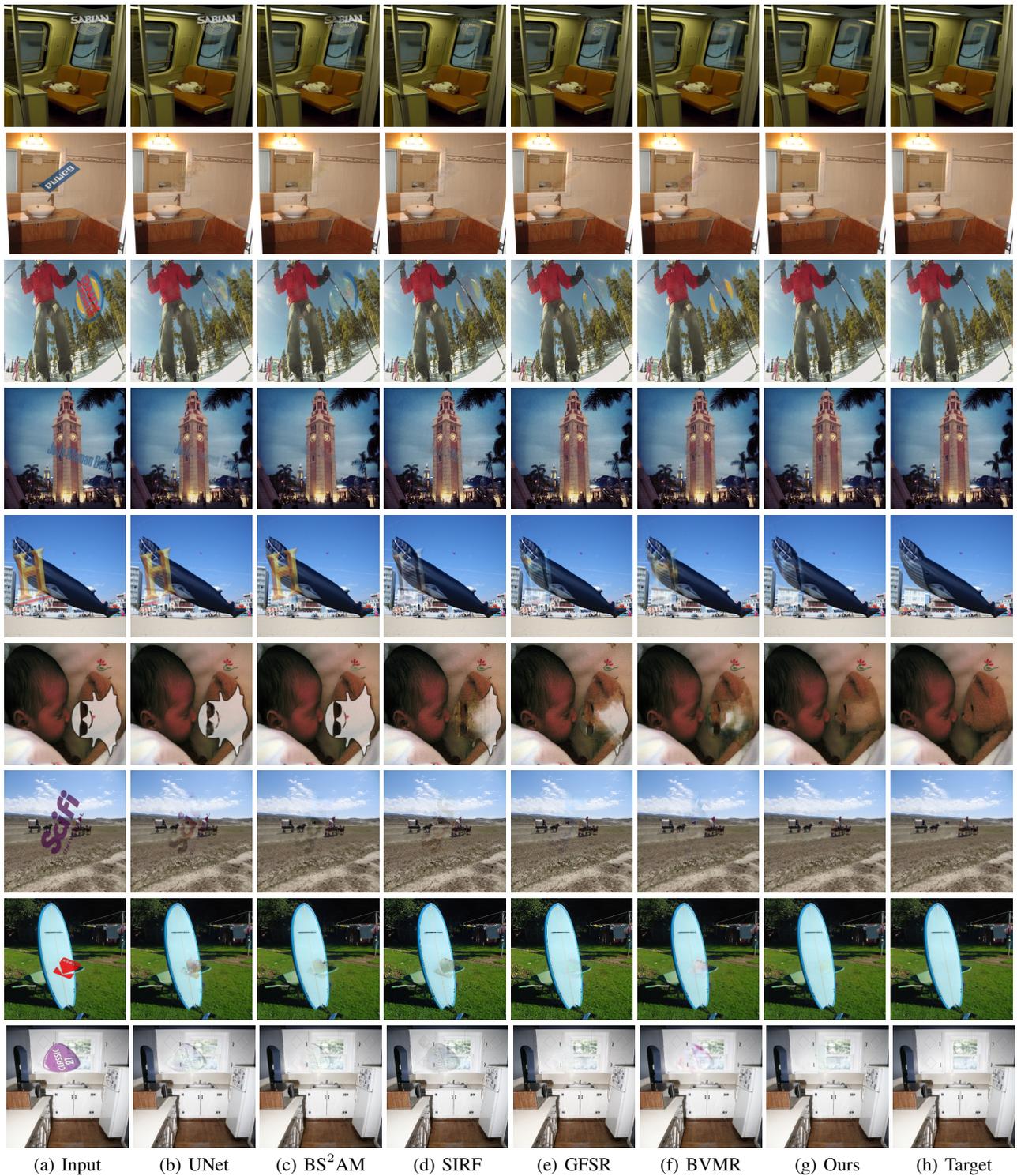

\centering     %%% not \center
\makeatletter
\@for\sun:=\smethods\do{\subfigure{\centering\includegraphics[width=\xwidth\textwidth]{\sima{\sun}}}\hspace{0.25em}}
\par\bigskip\vspace*{-2.em}

\@for\sun:=\smethods\do{\subfigure{\centering\includegraphics[width=\xwidth\textwidth]{\simi{\sun}}}\hspace{0.25em}}
\par\bigskip\vspace*{-2.em}

\@for\sun:=\smethods\do{\subfigure{\centering\includegraphics[width=\xwidth\textwidth]{\simc{\sun}}}\hspace{0.25em}}
\par\bigskip\vspace*{-2.em}

\@for\sun:=\smethods\do{\subfigure{\centering\includegraphics[width=\xwidth\textwidth]{\simb{\sun}}}\hspace{0.25em}}
\par\bigskip\vspace*{-2.em}

\@for\sun:=\smethods\do{\subfigure{\centering\includegraphics[width=\xwidth\textwidth]{\sime{\sun}}}\hspace{0.25em}}
\par\bigskip\vspace*{-2.em}

\@for\sun:=\smethods\do{\subfigure{\centering\includegraphics[width=\xwidth\textwidth]{\simf{\sun}}}\hspace{0.25em}}
\par\bigskip\vspace*{-2.em}

\@for\sun:=\smethods\do{\subfigure{\centering\includegraphics[width=\xwidth\textwidth]{\simg{\sun}}}\hspace{0.25em}}
\par\bigskip\vspace*{-2.em}

\@for\sun:=\smethods\do{\subfigure{\centering\includegraphics[width=\xwidth\textwidth]{\simh{\sun}}}\hspace{0.25em}}
\par\bigskip\vspace*{-2.em}

\makeatother

\addtocounter{subfigure}{-64}
\subfigure[Input]{\centering\includegraphics[width=\xwidth\textwidth]{\simk{input_256}}}
\subfigure[UNet]{\centering\includegraphics[width=\xwidth\textwidth]{\simk{unet}}}
\subfigure[BS$^2$AM]{\centering\includegraphics[width=\xwidth\textwidth]{\simk{urasc}}}
\subfigure[SIRF]{\centering\includegraphics[width=\xwidth\textwidth]{\simk{sirfimage}}}
\subfigure[GFSR]{\centering\includegraphics[width=\xwidth\textwidth]{\simk{gfsrimage}}}
\subfigure[BVMR]{\centering\includegraphics[width=\xwidth\textwidth]{\simk{bvmr}}}
\subfigure[Ours]{\centering\includegraphics[width=\xwidth\textwidth]{\simk{vvv4n}}}
\subfigure[Target]{\centering\includegraphics[width=\xwidth\textwidth]{\simk{target_256}}}

\caption{More comparisons between our method and other state-of-the-art methods over different proposed datasets. }
\label{fig:more}
\end{figure*}

\end{document}